\pdfoutput=1

\documentclass[preprint, twocolumn, 5p]{elsarticle}

\usepackage{amssymb}

\usepackage{algorithm}
\usepackage{algorithmic}
\usepackage {bm}
\usepackage{float}
\usepackage{booktabs}
\usepackage{multirow}
\usepackage{arydshln} 
\usepackage{array}
\usepackage{graphicx}  
\usepackage{wrapfig}  
\usepackage{amsmath}
\usepackage{subfigure}
\usepackage[pagebackref,breaklinks,colorlinks]{hyperref}
\usepackage{multicol} 
\usepackage{tabularx}

\journal{Neural Networks}

\begin{document}

\begin{frontmatter}



\title{DualFluidNet: an Attention-based Dual-pipeline Network for Fluid Simulation}


\author{Yu Chen}
\author{Shuai Zheng\corref{cor1}}
\cortext[cor1]{Corresponding author}
\ead{shuaizheng@xjtu.edu.cn}
\author{Menglong Jin}
\author{Yan Chang}
\author{Nianyi Wang}

\affiliation{organization={School of Software Engineering, Xi'an Jiaotong University},
            city={Xi'an},
            postcode={710049}, 
            country={China}}

\begin{abstract}
Fluid motion can be considered as a point cloud transformation when using the SPH method. Compared to traditional numerical analysis methods, using machine learning techniques to learn physics simulations can achieve near-accurate results, while significantly increasing efficiency. In this paper, we propose an innovative approach for 3D fluid simulations utilizing an Attention-based Dual-pipeline Network, which employs a dual-pipeline architecture, seamlessly integrated with an Attention-based Feature Fusion Module. Unlike previous methods, which often make difficult trade-offs between global fluid control and physical law constraints, we find a way to achieve a better balance between these two crucial aspects with a well-designed dual-pipeline approach. Additionally, we design a Type-aware Input Module to adaptively recognize particles of different types and perform feature fusion afterward, such that fluid-solid coupling issues can be better dealt with. Furthermore, we propose a new dataset, Tank3D, to further explore the network’s ability to handle more complicated scenes. The experiments demonstrate that our approach not only attains a quantitative enhancement in various metrics, surpassing the state-of-the-art methods, but also signifies a qualitative leap in neural network-based simulation by faithfully adhering to the physical laws. Code and video demonstrations are available at \href{https://github.com/chenyu-xjtu/DualFluidNet}{https://github.com/chenyu-xjtu/DualFluidNet}. 
\end{abstract}





\end{frontmatter}


\section{Introduction}
\label{sec:intro}

\begin{figure}
\centering
\subfigure[Real-time rendered fluid surface]{
		\includegraphics[width=\linewidth]{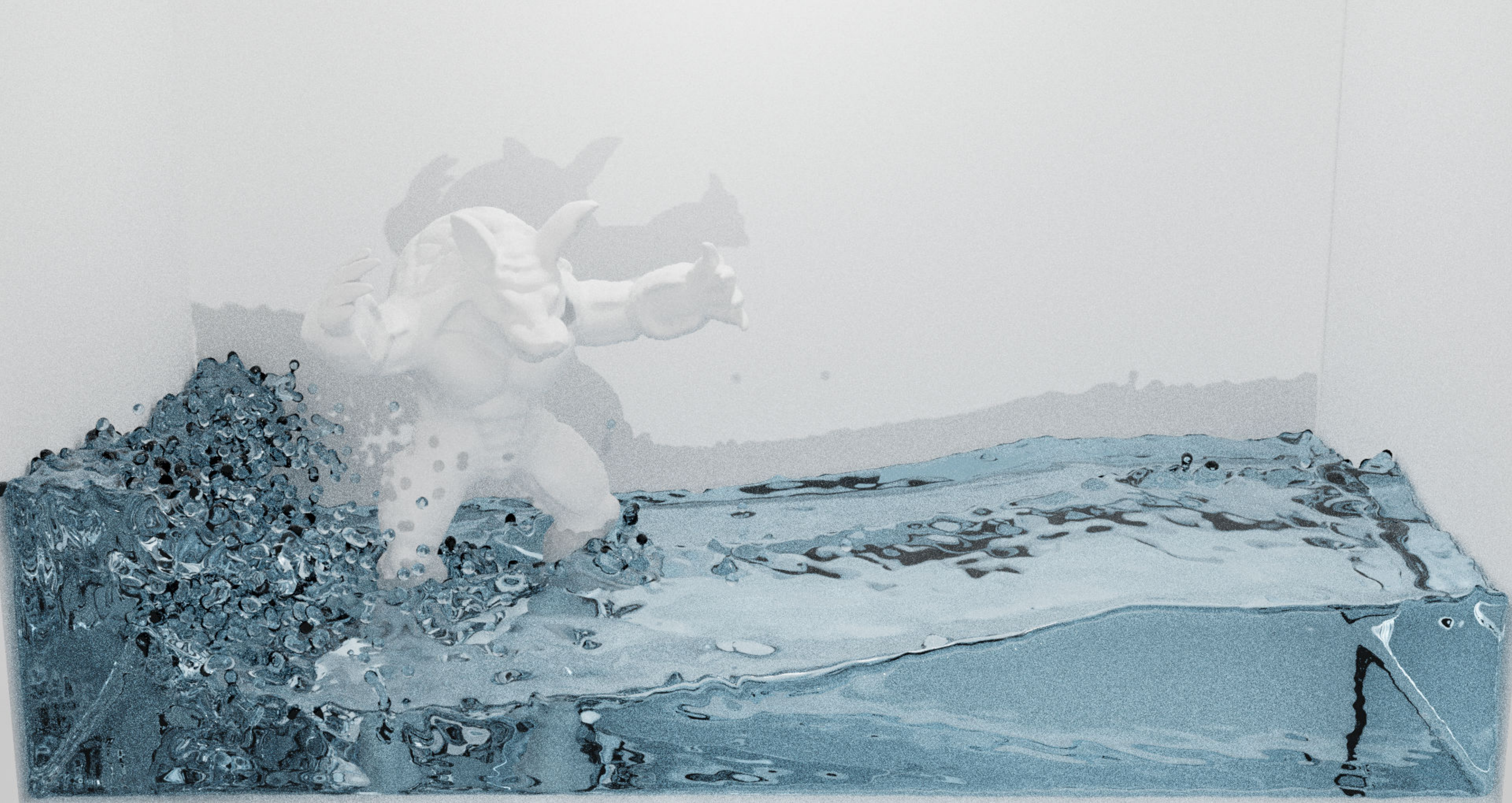}}\\
\subfigure[Underlying simulation particles]{
		\includegraphics[width=\linewidth]{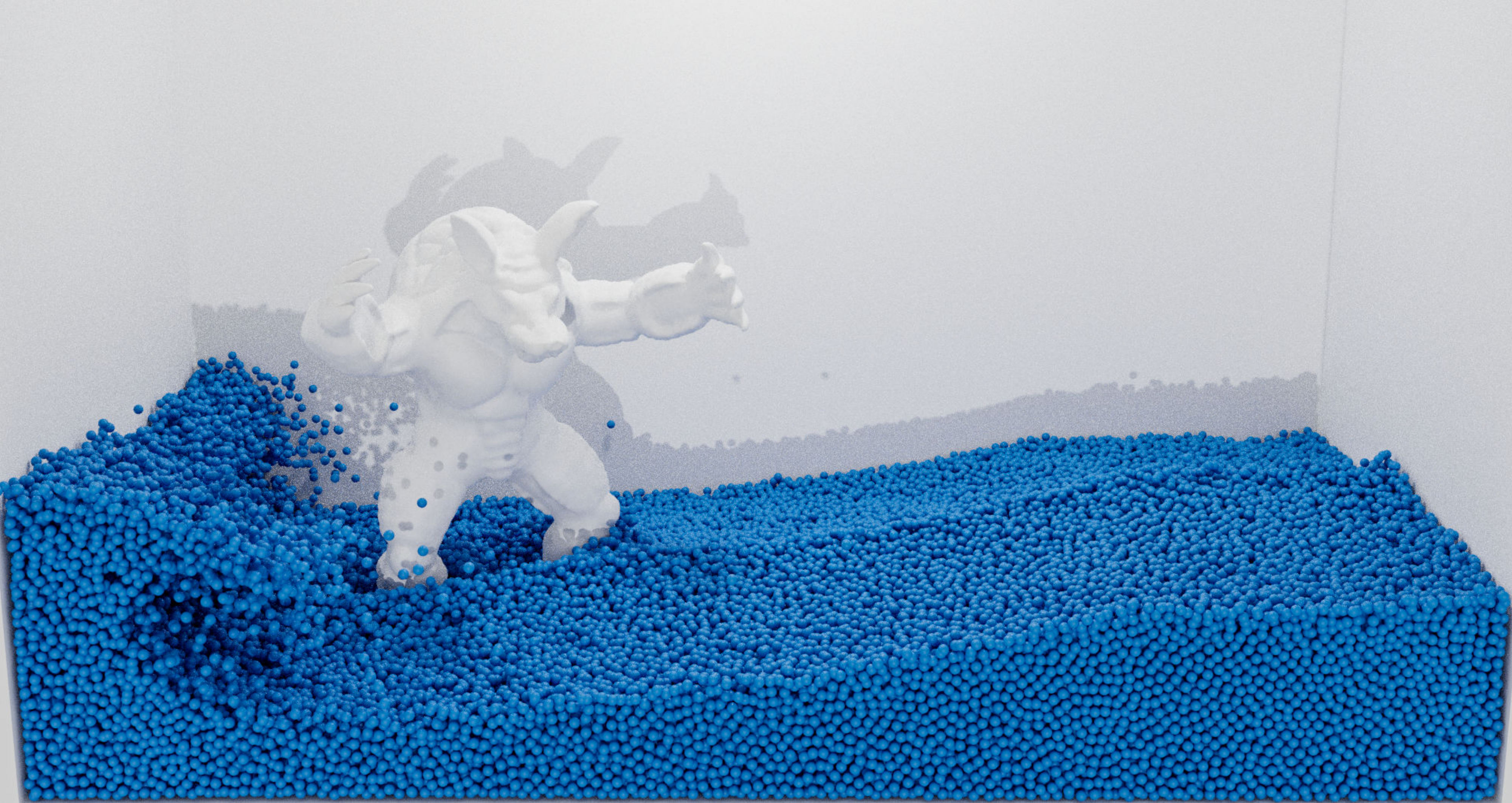}}
\caption{Armadillo in the flowing water, with 138k particles.}
    \label{fig:armadillo}
\end{figure}

Comprehending physics is essential for enhancing our understanding of the environment and our interactions with it. Traditional fluid simulations have limitations in practical applications due to their high computational demands. Recent studies \cite{JIN2021109951, woodward2023physics, cai2021physics,  KASHEFI202380, GAO202082,DENG2024106085} have highlighted the potential of physics-based neural networks as a promising approach for comprehending intricate natural phenomena from data. Inspired by Smoothed Particle Hydrodynamics (SPH) methods \cite{gingold1977smoothed,koschier2020smoothed}, which represent fluids as smoothed particles, several networks \cite{kakuda2021data, li2018learning, prantl2022guaranteed,sanchez2020learning,ummenhofer2019lagrangian, guan2022neurofluid} have been developed for particle-based simulations. The prediction of physical properties using neural networks is always treated as a regression problem\cite{ladicky2015data}, enabling the effective learning of physical phenomena. 

While previous methods have demonstrated excellent performance in fluid simulation by using neural network methods, we have observed that previous methods always struggle to maintain a balance between overall particle control accuracy and adherence to physical laws. The dilemma in previous methods turns into a trade-off issue. This limitation can result in deformations and disintegrations that defy the laws of physics, sometimes even in the simplest cases of free-fall motion. To deal with this problem, we propose a Dual-Pipeline Network architecture, consisting of a main pathway and a constraint-guided pathway. On one hand, the main pathway exhibits stronger learning capabilities of overall fluid dynamics, thereby being responsible for establishing a global context and stabilizing the fluid. On the other hand, although the constraint-guided pathway has relatively weaker overall learning capabilities, it can comprehend fluid particle motion while adhering to physical laws, thereby integrating physical laws into the main pathway. Our network seamlessly fuses the two learning ways through a meticulously designed approach, achieving a harmonious balance between these two pathways and unlocking their maximum potential. Instead of manually deciding how much each of the two pathways contributes, we propose a new module designed specifically for fluid particles, enabling each layer of the network to autonomously determine the optimal feature fusion parameters. This allows the network to automatically learn the best way to combine features from both pathways.

The integration of both fluid and solid inputs is often neglected in neural network-based methods for fluid simulation. Although previous networks appear to perform well in handling fluid point clouds, there hasn't been an effective method for handling the mixed input of fluid and solid. Common practices involve simply concatenating the two inputs and feeding them into the network \cite{ummenhofer2019lagrangian}, or passing each through a few convolutional layers before concatenation \cite{prantl2022guaranteed}. These straightforward input concatenations are insufficient to enable the network to effectively distinguish between the features of fluid and solid particles. Nevertheless, the fluid-solid coupling is crucial for enhancing fluid simulation performance. In the context of fluid simulation, our input comprises two entirely different features: fluid particles and solid particles. Additionally, the particle count varies in each input. Based on this observation, we propose an innovative Type-aware Input Module to facilitate better differentiation between fluid and solid particles in the input, enabling the network to more accurately and reliably compute their interactions during collisions. We verified that this module can be easily transferred to other network models and plays a crucial role in enhancing fluid simulation performance.

We conducted experiments on multiple 3D liquid datasets, and the results demonstrated that our network outperformed existing methods across various performance metrics, surpassing even the state-of-the-art network \cite{prantl2022guaranteed}. We also propose a new 3D liquid dataset, Tank3D, to further explore the network’s ability to handle more complex scenes. Furthermore, we validated the strong generalization of our network model in more challenging scenarios.

\section{Related work}
\label{sec:related}
\subsection{Representation of Fluid}
In Computational Fluid Dynamics (CFD), fluids are typically represented using either the grid-based or particle-based methods, the latter also known as Smoothed Particle Hydrodynamics (SPH). The grid-based method divides the fluid domain into a grid of cells or voxels. Properties such as velocity, pressure, and density are assigned to each grid point. The SPH method represents fluids as discrete particles. Each particle carries properties such as velocity, mass, and density. Properties at specific locations are computed  by averaging the attributes of neighboring particles. This process is performed using a kernel function, which assigns weights to nearby particles based on their distances from the target position. The SPH kernel functions produce non-zero values within a fixed radius range and are zero outside this range. Because grid-based methods often struggle with handling complex geometries and regions undergoing substantial deformation, and SPH can better utilize GPU parallel computing capabilities, the SPH approach has been widely adopted for fluid representation in recent years.

Traditional SPH methods exhibit a high level of simulation fidelity \cite{solenthaler2009predictive, bender2015divergence, ye2019smoothed, zheng2021topology, li2018multidisciplinary}, but they often entail complex computations and significant computational expenses, thereby constraining their utility in situations that require real-time predictions. In contrast, using machine learning to simulate the SPH method has emerged as a highly promising approach to capturing the intricacies of natural phenomena from data without incurring enormous computational costs \cite{ling2016reynolds, tompson2017accelerating, morton2018deep, SAHA2021359}. Due to the strong demand for generalization and feature extraction capabilities in fluid simulation, deep learning methods have been increasingly favored in recent years over traditional machine learning approaches \cite{ladicky2015data}. In utilizing neural networks for physics-based learning, a crucial consideration is selecting the appropriate data representation. Point clouds are widely employed in computer vision and robotics applications to represent the 3D environment \cite{qi2017pointnet, qi2017pointnet++, wang2019dynamic, chen2023rotation}. In the SPH method using neural networks, fluid particles can also be regarded as point clouds, while fluid simulation can be visualized as point cloud transformation.

\subsection{Neural Network Approaches in Fluid Simulation}
In time series analysis and prediction tasks, RNN-based models such as Long Short-Term Memory (LSTM) or Gated Recurrent Unit (GRU) are commonly employed \cite{Learning-spatiotemporal, LSTM1}. This is attributed to the temporal dependencies and sequential nature inherent in time series data, which are effectively captured by recurrent neural network models. While using RNN-based networks to predict the state of fluid particles frame by frame might seem like an intuitive choice, we observe that employing CNN-based networks is more appropriate in practice. By using convolution kernel to simulate the kernel function in SPH method, the state (position and velocity) of fluid particles at timestep \(t+1\) can be directly computed from the particle state at timestep \(t\), without requiring information from timesteps prior to \(t\). Conversely, incorporating the hidden states of RNN-based networks will introduce instability into the computation.

A common convolution approach is to utilize graph structures \cite{shao2022transformer, li2018learning, battaglia2016interaction, sanchez2020learning, mrowca2018flexible} to create a dynamic graph connecting nearby particles, facilitating the modeling of fluid behavior. This is achieved by updating nodes and edges to simulate message passing between SPH particles. Graph-based methods are tied to particle discretization, but a wide range of physical phenomena, such as fluid mechanics, are governed by continuous partial differential equations (PDE) instead of discrete graph structures. Therefore, continuous methods \cite{Wang_2018_CVPR, ummenhofer2019lagrangian, prantl2022guaranteed,thomas2019kpconv} always exhibit a stronger capability. Most related to our approach is \cite{ummenhofer2019lagrangian, prantl2022guaranteed}. \cite{ummenhofer2019lagrangian} proposed a continuous convolution kernel (CConv) designed for establishing connections between particles and their neighbors. CConv exhibits strong learning capabilities for fluid particles and can broadly capture the dynamics of fluid motion. However, due to the absence of explicit physical constraints, its predictions may deviate from physical laws. Therefore, \cite{prantl2022guaranteed} has designed an anti-symmetric continuous convolution kernel (ASCC) building upon the CConv to introduce a robust constraint, allowing the network to approximate the conservation of momentum in physics during training. However, this strong constraint also complicates the problem for the network, leading to a reduction in its learning capacity. These networks always face such a dilemma, while our dual-pipeline network effectively integrates the strengths of both distinct learning approaches, enabling our network to achieve a better equilibrium between global fluid control and adherence to physical law constraints.

\subsection{Feature Fusion}
Common feature fusion methods often appear in problems related to multimodal fusion \cite{zhao2023cddfuse, liu2023sfusion}, multi-view image fusion \cite{Variational-gated, Fusion1}, multiscale feature fusion \cite{Fusion-Based, Context-Aware, dai2021attentional, Underwater}, and so on. Different from the scenarios mentioned above, feature fusion in this paper is addressed in two scenarios: the fusion of features from two pathways (Attention-based Feature Fusion), and the blending of fluid particles and solid particles inputs (Type-aware Input Module). As for the Attention-based Feature Fusion, the emphasis is on leveraging the strong learning capability of the main path to capture global fluid features, while using the constraint-guided path to incorporate physical information into the main path. Compared to a single pathway, the fusion of two pathways complements each other and maximizes the advantages of both. And the Type-aware Input Module emphasizes a better understanding and differentiation between two distinct types of input: fluid and solid particles in the input. Its purpose is to enable the network to accurately and reliably compute their interactions during collisions.

Both the Attention-based Feature Fusion and Type-aware Input Module utilize a similar soft-attention mechanism, specifically designed for fluid particles. We also consider other attention mechanisms, particularly self-attention and cross-attention \cite{vaswani2017attention}. Both mechanisms operate by computing the dot product between queries (Q), keys (K), and values (V), which are transformation representations of elements in the sequence. Self-attention is especially effective for capturing long-range dependencies within a sequence, while cross-attention helps the model capture alignment relationships between two sequences. These attention mechanisms are also commonly employed in feature fusion tasks, as they enable the model to directly compute interactions between any two positions in the sequence. However, in SPH fluid simulation, the local support property of SPH kernel functions ensures that they only produce non-zero values within a fixed radius range, being zero outside of this range. This implies that each particle is influenced only by nearby particles within this radius, with far particles having little effect. Consequently, SPH simulations do not require self-attention and cross-attention mechanisms to explicitly capture long-range dependencies. Instead, through our designed CConv-based soft-attention feature fusion for fluid simulation, we can better fit the local properties of the kernel function, effectively simulating particle interactions while ensuring higher efficiency.
\section{Method}
\label{sec:method}
\begin{figure*}
  \centering
    \includegraphics[width=\linewidth]{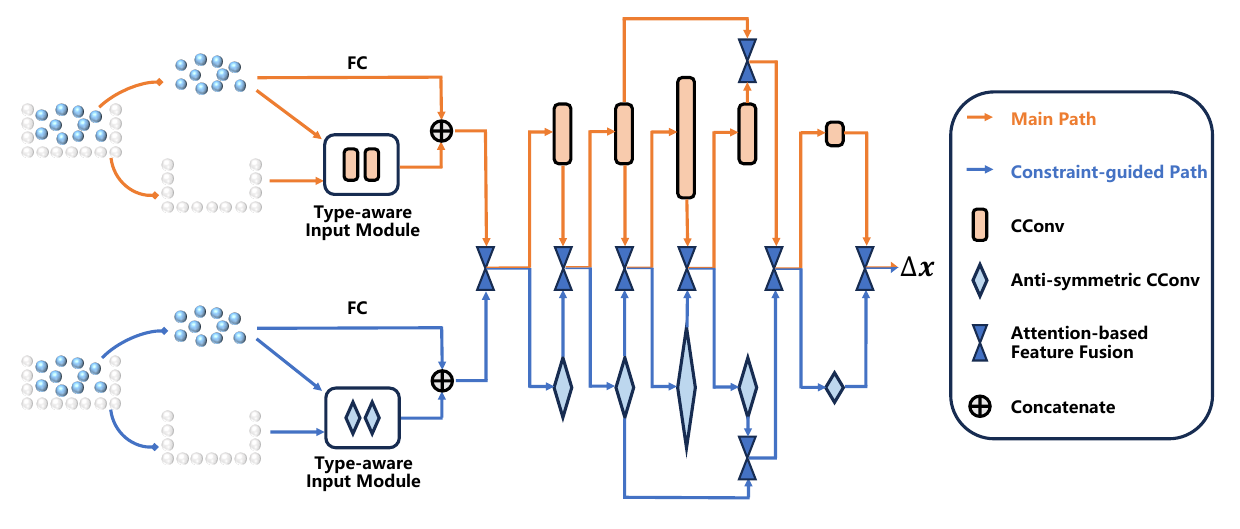}
    \caption{The architecture of our network. It consists of two pathways: the Main Path and the Constraint-Guided Path. These two pathways share a similar structure, with the Main Path utilizing CConv as its convolutional kernel (the orange rectangle), and the Constraint-Guided Path employing ASCC as its kernel (the blue diamond). At each layer, both pathways pass through a module for feature fusion before feeding the fused features into the next layer.}
    \label{fig:network}
\end{figure*}
Before introducing our method, the mathematical notation and preliminary definitions are provided in Table \ref{notations table}.
\begin{table}[htbp]
  \centering
  \begin{tabularx}{1\linewidth}{lX}
  \toprule
  \(F_{ext}\) & external force \\ 
  \(N_{\theta}\) & neural network with parameters \(\theta\)\\
    \(P_n\) & a set of \(n\) particles\\
    \(x_{i}^{t}, v_{i}^{t}\) & state of particle \(i\) at time \(t\) \\
    \(x^{t}, v^{t}\) & state of particles at time \(t\) \\
    \(x_{i}^{t*}, v_{i}^{t*}\) & state of particle \(i\) at time \(t\) after \(F_{ext}\) \\
     \(x^{t*}, v^{t*}\) & state of particles at time \(t\) after \(F_{ext}\) \\
     \(\Delta x^{t+1}\) & displacement of particles by internal forces \\
    \(x^{t+1}, v^{t+1}\) & state of predicted particles at time \(t+1\) \\
    \(\hat{x}_{i}^{t+1}, \hat{v}_{i}^{t+1}\) & state of ground-truth particles at time \(t+1\) \\
    \midrule
    \(*\) & convolution operator \\
    \( \Lambda\) & a mapping function that transforms a unit ball into a unit cube\\
    \(\left \| \ \cdot \ \right \|_{2}\) & \(L2\) norm (or called Euclidean norm) \\
    \(\phi\) & function of \textit{CConv} \\
    \(\psi\) & function of \textit{ASCC} \\
    \(\oplus\) & concatenate\\ 
    \(\otimes\) & weighted average based on weight \(w\)\\
    \(S\) & function of \textit{Particle Selector} \\
    \bottomrule
  \end{tabularx}
  \caption{A summary table of the main notations. The "state of particle" includes the position and velocity.}
  \label{notations table}
\end{table}

\begin{algorithm}
\renewcommand{\algorithmicrequire}{\textbf{Input:}}
\renewcommand{\algorithmicensure}{\textbf{Output:}}
\caption{Simulation Process}
	\label{alg1} 
	\begin{algorithmic}
        \STATE  $\bm{Input:} x^t=\left \{x^t_{1}, x^t_{2}, ... x^t_{n}\right \}, v^t=\left \{v^t_{1}, v^t_{2}, ... v^t_{n}\right\}$
        \FORALL{particles i}
            \STATE apply external forces $v_{i}^{t*} = v^t_i + \Delta t\frac{F_{ext}}{m}$\;
            \STATE update position $x_{i}^{t*} = x_{i}^t + \Delta t v_{i}^{t*}$\;
        \ENDFOR
        \STATE $x^{t*} = \left \{x_{1}^{t*}, x_{2}^{t*}, ... x_{n}^{t*}\right \}$
        \STATE $v^{t*} = \left \{v_{1}^{t*}, v_{2}^{t*}, ... v_{n}^{t*}\right \}$
		\STATE predict position $\Delta x^{t+1}=N_{\theta}\left (x^{t*}, v^{t*}\right )$ 
		\STATE update position $x^{t+1}=x^{t*} + \Delta x^{t+1}$
        \STATE update velocity $v^{t+1}=\frac{x^{t+1} - x^t }{\Delta t}$
        \STATE $\bm{Output:} x^{t+1}, v^{t+1}$
	\end{algorithmic} 
\end{algorithm}

Our network follows the position-based fluids (PBF) \cite{macklin2013position} scheme. Algorithm \ref{alg1} outlines the general procedure of this approach, where \(N_{\theta}\) represents our neural network model with trainable parameters \(\theta\), and \(\Delta t\) represents the time interval between two consecutive time steps. Given a set of particles, denoted as \(P_n\), where \(n\) represents the particle number, the algorithm proceeds to analyze their positions and velocities. It takes \(x^t\) and \(v^t\) as inputs, which represent the positions and velocities of the particles at time step \(t\). Our central learning goal is to approximate the underlying physical dynamics to predict the positions and velocities of particles at the next time step \(t+1\).

\subsection{Two Types of Convolution Kernels}
We introduce two types of convolution kernels, denoted as CConv and ASCC. Given a point cloud \(P_n\) containing \(n\) points labeled as \(i=1, ..., n\), with corresponding values \(f_i\) at positions \(x_i\), the convolution of CConv at position \(x \in P_n\) can be defined as follows:
\begin{equation}\label{eq2}
    \begin{aligned}
CConv_{g}&=\left ( f*g \right )\left ( x \right ) \\   &=\sum_{i\in \mathcal{N}\left (x,R  \right )}^{}a\left (x_{i}, x  \right )f_{i}g\left ( \Lambda \left ( x_{i} -x \right )  \right ).
    \end{aligned}
\end{equation}
Where \(\mathcal{N}\left (x, R \right )\) represents the set of points located within a radius R around the position x. \(g\) is the filter function, utilizing a mapping \(\Lambda (r)\) that transforms a unit ball into a unit cube, described by \cite{griepentrog2008bi, ummenhofer2019lagrangian}. \(a\) is a window function applied for density normalization \cite{hermosilla2018monte} specific to the points \(x_i\) and \(x\):
\begin{equation}\label{eq3}
a\left ( x_{i},x  \right ) =\left\{ 
    \begin{array}{lc}
        \left ( 1-\frac{\left \| x_{i}-x \right \|_{2}^{2} }{R^{2} }  \right ) ^{3}  \quad  & \left \| x_{i}-x \right \|_{2} < R  
  \\
         0&else.\\
    \end{array}
\right.
\end{equation}
\begin{wrapfigure}{r}{4cm}  
		\centering
		\includegraphics[width=1\linewidth]{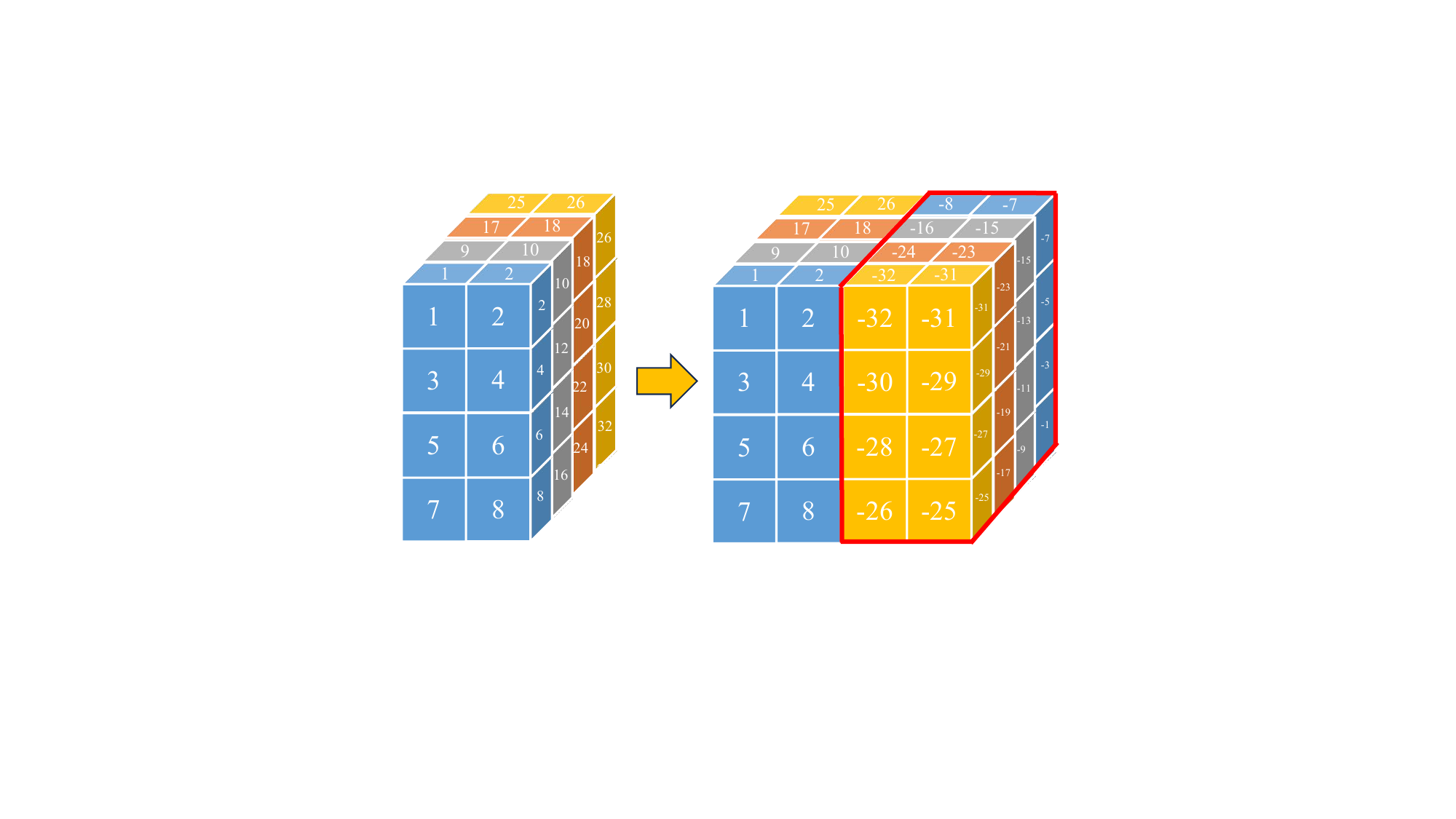}
		\caption{Anti-symmetric kernel is obtained by negating and mirroring the trainable variables by the center point.}
		\label{antisymmetric}
\end{wrapfigure}

ASCC can be regarded as a specialized variant of CConv, where it enforces a strong constraint through the antisymmetric kernel design, allowing it to learn fluid features while adhering to the conservation of momentum. This is achieved by halving the learnable kernel parameters along a chosen axis and determining the second half through reflection about the center of the kernel, shown in Figure \ref{antisymmetric}. Notably, the mirrored values are negated. ASCC can be defined as:
\begin{equation}\label{eq4}
    \begin{aligned}
ASCC_{g_{s}}&=\left ( f*g_s \right )\left ( x \right ) \\ &=\sum_{i\in \mathcal{N}\left (x,R  \right )}^{}a\left (x_{i}, x  \right )(f+f_{i})g_s\left ( \Lambda \left ( x_{i} -x \right )  \right ).
    \end{aligned}
\end{equation}
Where \(g_s\) represents the anti-symmetric continuous convolution kernel and \(f\) represents the corresponding value at the position \(x\), with the remaining symbols defined similarly to the definitions in Equation \ref{eq2}. The antisymmetric kernel \(g_s\) satisfies the constraint condition \(g_s(x)=-g_s(-x)\). For two distinct points \(x,y\in P_n\), the interparticle force between these two points can be formulated as:
\begin{equation}\label{eq:Fxy=Fyx}
    \begin{aligned}
    F_{xy}&=(f_x+f_y)(-g_s(x-y)) \\
         &=-F_{yx}.
    \end{aligned}
\end{equation}

Then Equation \ref{eq:Fxy=Fyx} can further lead to the validity of the following equation:
\begin{equation}
    \int_{x\in P_n}^{} \int_{y\in P_n}^{} F_{xy}=0.
\end{equation}

The antisymmetric design of ASCC ensures compliance with momentum conservation laws in data-driven learning, but this strong constraint also reduces its learning capacity, making the problem much harder to solve. In contrast, CConv has better fitting and learning capabilities but lacks constraints of physical laws. In the next section, we introduce our proposed Dual-pipeline architecture, which aims to build a neural network that integrates strong learning abilities while strictly adhering to physical principles.

\begin{figure}
  \centering
  \subfigure[Attention-based Feature Fusion]{
    \label{fig:attention_based_feature_fusion} 
    \includegraphics[width=0.52\linewidth]{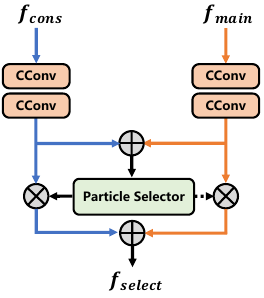}}
  \subfigure[Particle Selector]{
    \label{fig:Particle Selector} 
    \includegraphics[width=0.42\linewidth]{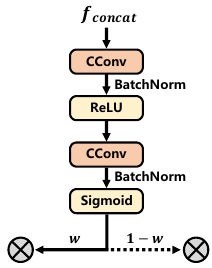} }

  \caption{The architecture of the Attention-based Feature Fusion Module. Features from both pathways pass through two layers of CConv and are then concatenated. The concatenated features are subsequently input into a Particle Selector to determine the respective weights for fusion. The original features from both pathways are then combined with weighted fusion.} 
\label{fig:Attention-based Module}
\end{figure}

\subsection{Dual-pipeline Architectrue}
Our network architecture is illustrated in Figure \ref{fig:network}. Instead of subjectively determining the relative significance of the two pathways, we employ an Attention-based Feature Fusion Module, allowing the network to autonomously learn the optimal connection between the two pathways, shown as Figure \ref{fig:attention_based_feature_fusion}.

More precisely, our network comprises a main pathway and a constraint-guided pathway. Both pathways share a similar structure, each comprising a Type-aware Input Module and a sequence of continuous convolutions with an effective depth of five. The main pathway, which utilizes CConv as its kernel, is responsible for establishing a global context and stabilizing the fluid. The Constraint-guided pathway employs ASCC as its kernel, which enforces a strong constraint to ensure momentum conservation while learning fluid motion patterns. Before feeding the input into the convolutional sequence, we combine the result from the input module and a fully-connected layers by addition. In the second and fourth layers of the convolution sequence, we leverage the Feature Fusion Module to implement a residual connection concept \cite{he2016deep}. We found that these designs improve accuracy and robustness. 

\begin{figure}
    \centering
    \includegraphics[width=0.9\linewidth]{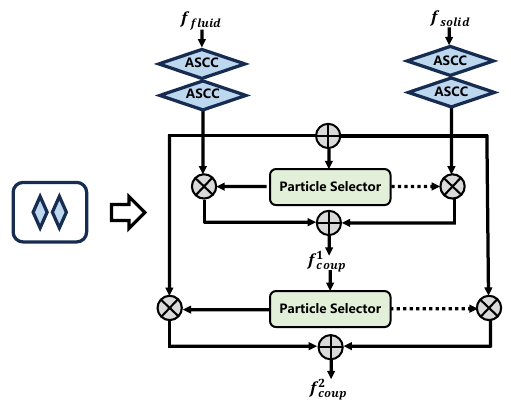}
    \caption{The architecture of the Type-aware Input Module in the Constraint-guided Path, with ASCC as the kernel. In contrast, the Main Path's Type-aware Input Module replaces the ASCC kernel with CConv. The Particle Selector is the same as Figure \ref{fig:Particle Selector}.} 
    \label{fig:Type aware}
\end{figure}

Our network seamlessly fuses these two pathways through an Attention-based Feature Fusion Module to achieve a harmonious balance between the two pathways and unlock their maximum potential. In each layer of the convolutional sequence, assuming the features from the Main Path as \(f_{main}\) and the features from the Constraint-guided Path as \(f_{cons}\), the fusion of these two features in the Feature Fusion Module can be represented as follows:
\begin{equation}\label{eq5}
    \begin{aligned}
f_{fusion}&=S(\phi (f_{main})\oplus \phi (f_{cons})) \otimes f_{cons}\\
&+(1-S(\phi (f_{main})\oplus \phi (f_{cons}))) \otimes f_{main}.
    \end{aligned}
\end{equation}
Where \(\phi \) represents the \textit{CConv} function(Equation \ref{eq2}). \(S\) represents the Particle Selector and its architecture is illustrated in Figure \ref{fig:Attention-based Module}. With the concatenated features from the two pathways as input, the selector outputs a selection weight between 0 and 1, which is used to assess the respective significance of features from both pathways for fusion, aiming to achieve an optimal feature fusion. Unlike previous approaches, this kind of Attention-based fusion method allows our network to learn the optimal fusion between the two pathways autonomously, endowing our network with strong learning capabilities while ensuring compliance with the law of momentum conservation.

\subsection{Type-aware Input Module (TaIM)}
We propose an approach to facilitate better differentiation between fluid and solid particles in the input, promoting a better understanding of fluid-solid coupling. Consequently, this enables the network to more accurately and reliably compute their interactions during collisions. Figure \ref{fig:Type aware} provides an example of the Type-aware Input Module architecture in the Constraint-guided Path, with ASCC as the kernel. In contrast, the Type-aware Input Module in the Main Path replaces the ASCC kernel with the CConv kernel. Unlike the Attention-based Feature Fusion Module between two pathways, the input features here represent two entirely different types, posing a greater challenge for our network to recognize their respective diverse properties. The quality of the initial fusion can significantly influence the final fusion weights. Since it is still a feature fusion problem, an intuitive approach is to employ an additional attention module for fusing these input features, which can be represented as:
\begin{equation}\label{eq6}
    \begin{aligned}
f_{coup}^{1}&=S(\psi  (f_{fluid})\oplus \psi (f_{solid})) \otimes f_{fluid}\\
&+(1-S(\psi (f_{fluid})\oplus \psi (f_{solid}))) \otimes f_{solid}.
    \end{aligned}
\end{equation}

\begin{equation}\label{eq7}
f_{coup}^{2}=S(f_{coup}^{1}) \otimes f_{fluid}+(1-S(f_{coup}^{1})) \otimes f_{solid}.
\end{equation}
Where \(\psi\) represents the \textit{ASCC} function (Equation \ref{eq4}). \(f_{coup}^{1}\) represents the output of the first attention module, while \(f_{coup}^{2}\) serves as the module's final output. 

\subsection{Training Procedure}
Consistent with prior research \cite{ummenhofer2019lagrangian,li2018learning,sanchez2020learning}, we employ a loss function that calculates the mean absolute error in position values between the predicted and ground truth weighted by the neighbor count as our loss function:
\begin{equation}\label{eq8}
\mathcal L^{t+1} = \sum_{i=1}^{N}\varphi_i\left \| x_{i}^{t+1} - \hat{x}_{i}^{t+1}  \right \|_{2}^{\gamma }.
\end{equation}
Where \(x_{i}^{t+1}\) represents the predicted position from the Algorithm \ref{alg1} at time step \(t+1\), and \(\hat{x}_{i}^{t+1}\) represents the ground-truth position. \(\varphi_i\) is the weight of neighbor count to emphasize the loss for
particles with fewer neighbors:
\begin{equation}\label{eq9}
\varphi_i=exp(-\frac{1}{c} \left |\mathcal{N}(x_{i}^{t*})\right |).
\end{equation}
\(\mathcal{N}(x_{i}^{t*})\) represents the neighbor count of the particle \(i\) at time step \(t\) after experiencing external force. We set \(c=40\), in accordance with the average number of neighboring particles and \(\gamma =0.5\) following \cite{ummenhofer2019lagrangian}, which makes the loss function more sensitive to small particle motions. For temporal stability considerations, we conduct training to predict particle positions for two future time steps, \(t+1\) and \(t+2\). This extension contributes to an overall enhancement in simulation quality. As a result, the combined loss function can be represented as follows:
\begin{equation}\label{eq10}
\mathcal L = \mathcal{L}^{t+1} + \mathcal{L}^{t+2}.
\end{equation}

\section{Experiments}
\label{sec:experiments}
Our experiments aimed to showcase the particle prediction accuracy, probability distribution divergence, adherence of physical law, and inference time of our approach compared to various baseline methods. What's more, we show the great generalization and broad applicability of our network with adequate visual analysis.

\subsection{Experimental Setup}
We utilize the PyTorch framework and employ the Adam optimizer for training, with the batch size set to 16. The initial learning rate is 0.002, and we halve the learning rate at steps 10000, 20000, ..., 50000. In total, we train our networks for 50,000 iterations on an NVIDIA RTX 3090Ti. Most other parameters follow the settings in \cite{ummenhofer2019lagrangian}. The particle radius is set to \(h=0.025m\), and the spherical filters with a spatial resolution of [4,4,4] are used with a radius of \(R=4.5h\).

\begin{table*}
    \renewcommand{\arraystretch}{}
  \centering
\resizebox{1.0\linewidth}{!}{
\scriptsize
\begin{tabular}{clcccccc}
    \toprule[0.6pt]
& {\multirow{2}{*}{Method}} & \multicolumn{2}{c}{Average pos error(mm)} & Average distance of & Wasserstein  & Max density & Frame inference\\ \cline{3-4}
                      &  & t+1            & {t+2} &\(n\) frames sequence \(d^n\)(mm) &   distance(mm)  &error(\(g/cm^{3}\)) &time(ms)                                 \\
                      \specialrule{0em}{0pt}{1pt}     \midrule \midrule
              {\multirow{5}{*}{\rotatebox{90}{Liquid(6k)}}}
         &  DFSPH \cite{bender2015divergence}    &      -      &     -        &   -  &       -    &-    &  \( \geq1,000  \)                                      \\
         &  DPI-Nets \cite{li2018learning}    &          26.01            &        50.67        &   unstable  &       -             &unstable    &    305.55                                       \\
         & KPConv \cite{thomas2019kpconv}          &       1.65               &       4.54         &  unstable    &          -       &unstable       &     57.89                                      \\
          &  PCNN \cite{Wang_2018_CVPR}    &          0.64            &        1.87        &   32.50  &      0.33    &0.15       &    187.34                                       \\
          &  Regression-Forests \cite{ladicky2015data}    &       0.68         &       1.97     &      38.47       &   0.29  & 0.13  & \textbf{12.98}                 \\ 
        &   CConv \cite{ummenhofer2019lagrangian}       &       0.60               &      1.55         &30.09    &      0.26           &0.12       &      16.47                                \\ 
           
        &   DMCF \cite{prantl2022guaranteed}          &       0.65               &      1.89          & 32.89    &      0.21          &0.07       &      94.86                                  \\ 
           
        &  \textbf{Ours}          &       \textbf{0.43}               &      \textbf{1.16}          &\textbf{28.32}    &      \textbf{0.17}      & \textbf{0.06}           &   48.01                                     \\ 
     \midrule
          {\multirow{5}{*}{\rotatebox{90}{DamBreak}}}
       &  DFSPH \cite{bender2015divergence}    &      -      &     -        &   -  &       -   &  -   &  \( \geq1,000  \)                                      \\
        &     DPI-Nets \cite{li2018learning}   &          12.70            &        24.38        &   22.57  &       -       &0.16           &    202.56         \\
       &   KPConv \cite{thomas2019kpconv}         &       2.49              &      7.05        &  unstable    &          -         &unstable     &    47.96                                     \\
       &  PCNN \cite{Wang_2018_CVPR}    &          0.75            &        1.93        &   25.43  &       0.31   &0.14    &    187.34    
       \\
         &  Regression-Forests \cite{ladicky2015data}    &       0.72         &       1.62     &      31.40         &   0.25  &0.14  & \textbf{11.10}                                     \\ 
         &  CConv \cite{ummenhofer2019lagrangian}         &       0.57               &      1.38       & 20.63    &      0.22       &0.11       &      12.01                            \\ 
           
         &  DMCF \cite{prantl2022guaranteed}         &       0.64               &      1.83          &24.56    &      0.20         &0.07    &       70.49                          \\ 
           
         &  \textbf{Ours}          &       \textbf{0.42}               &      \textbf{1.02}          &\textbf{20.09}   &      \textbf{0.18}         & \textbf{0.06}      &   35.57                                         \\ 
     \bottomrule[0.6pt]
\end{tabular}}
  \caption{Quantitative Evaluation. We evaluated the networks according to the evaluation metrics described in Section \ref{subsec:Evaluation metrics} and also compared the inference time required per frame for each network. It's worth noting that some methods become unstable after a few frames.}
  \label{tab:Quantitative Experiments}
\end{table*}
\subsection{Datasets}
To evaluate the fundamental fluid simulation capability, we conducted experiments on three basic 3D fluid datasets. Two of these datasets were derived from the Liquid3D dataset presented by \cite{ummenhofer2019lagrangian}. The first dataset, denoted as 'Liquid(complex),' encompasses a variety of complex scenarios featuring different initial shapes of fluid particles and multiple complex boxes, with particle counts ranging from thousands to tens of thousands. The second dataset, denoted as 'Liquid(6k),' represents a simpler scenario with a fixed particle count of 6000. The Liquid3D dataset was generated using DFSPH \cite{bender2015divergence}, which emphasizes high-fidelity simulations and operates at a 50 Hz time resolution. It comprises 200 training scenes and 20 test scenes. The third dataset, 'DamBreak,' was from \cite{li2018learning} and consists of 2000 training scenes and 300 testing scenes. These scenes simulate the behavior of a randomly positioned fluid block in a static box. We validate and compare the fundamental fluid simulation capabilities of all methods across these three datasets.

\begin{wrapfigure}{r}{4cm}  
		\centering
		\includegraphics[width=1\linewidth]{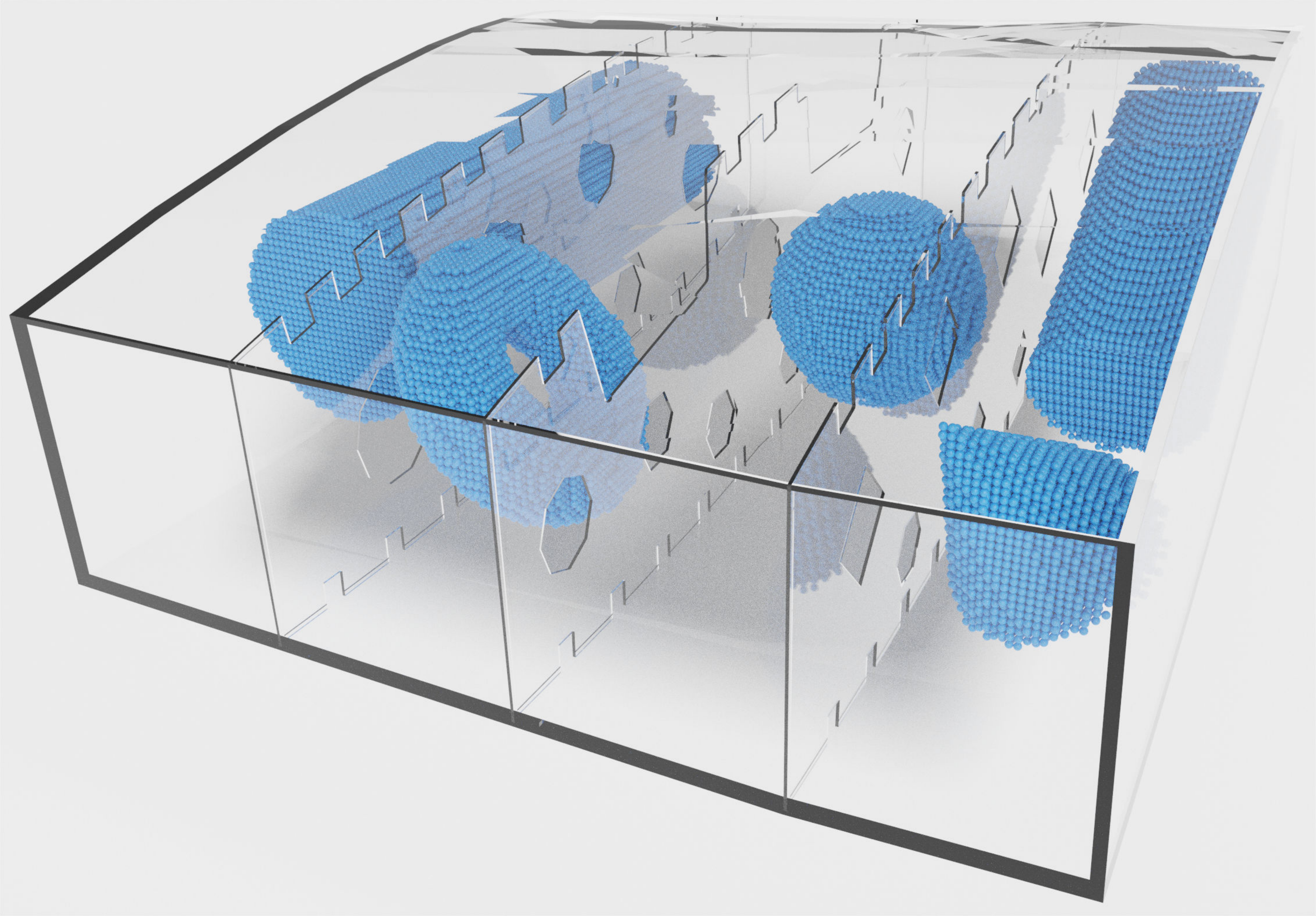}
		\caption{An example scene of the Tank3D dataset}
		\label{Tank3D}
\end{wrapfigure}
To further explore the network's ability to handle more complicated scenes, we propose a new dataset, Tank3D. Tank3D is constructed following the same methodology as Liquid3D \cite{ummenhofer2019lagrangian}, generated using DFSPH \cite{bender2015divergence}. It consists of 100 randomly generated training scenes and 10 test scenarios each featuring a tank with a complex shape comprising three baffles with numerous holes. In each scene, 4 to 16 fluid particles are randomly generated for free fall. Optional fluid shapes remain consistent with Liquid3D, described in \cite{ummenhofer2019lagrangian}, but undergo random scaling and rotation. An example scene is shown in Figure \ref{Tank3D}. This Tank3D dataset features significantly more complex, serving as an effective benchmark for evaluating the neural network's adherence to physical laws, fluid-solid interaction capability, and generalization ability. In Section \ref{complex capability}, we further investigate the network's handling abilities in complex terrains using the Tank3D dataset.

\begin{figure*}
  \centering
    \includegraphics[width=\linewidth]{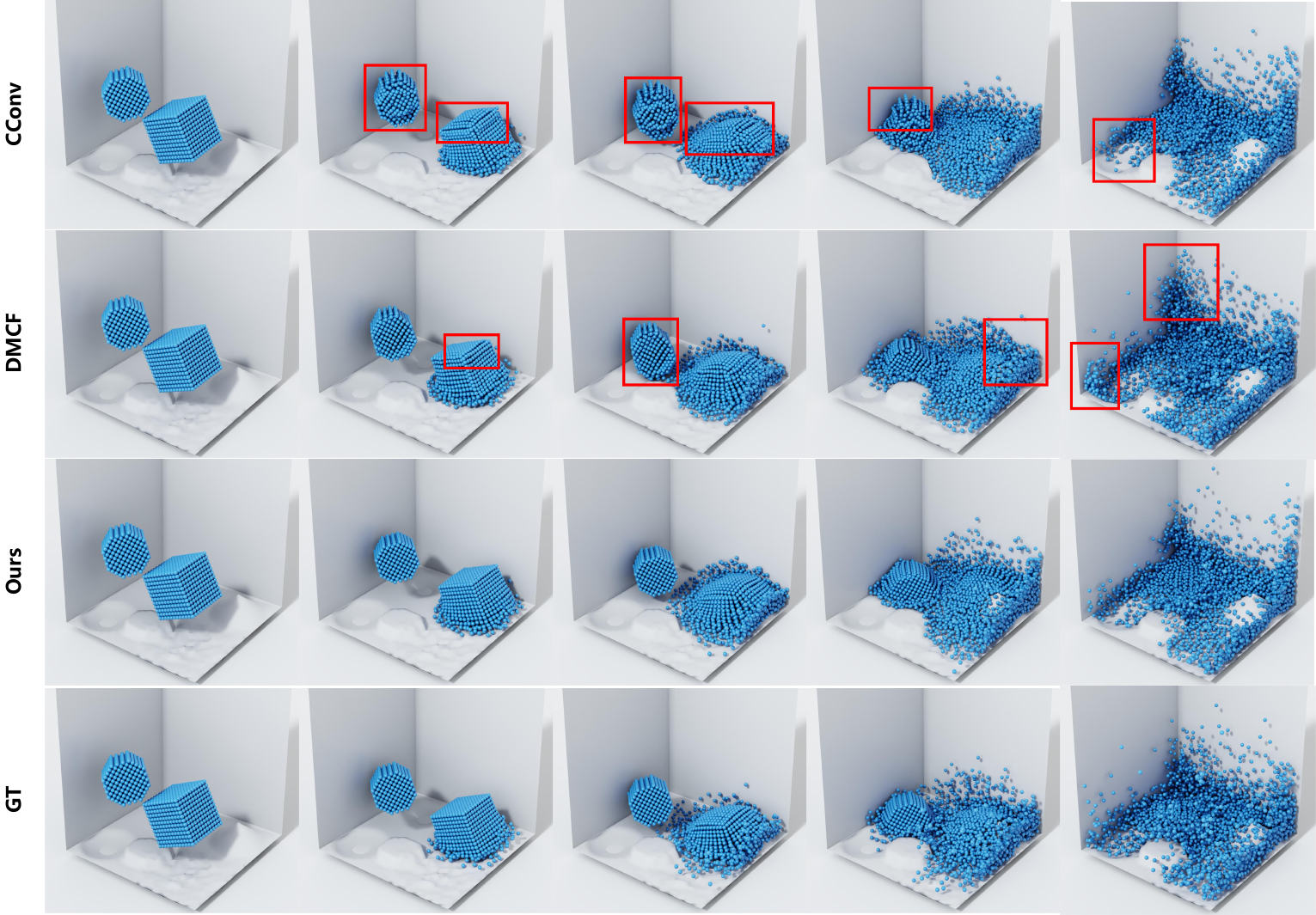}
    \caption{Qualitative Evaluation. In the results of CConv, the particles exhibit a clear overfitting to gravity. DMCF can effectively capture the physics of falling, but its ability to learn stable overall fluid dynamics is somewhat lacking, leading to occasional distortions at the edges. Our network better adheres to physical laws and reduces simulation errors, bringing it closer to the ground truth.}
    \label{fig:Qualitative Evaluation}
\end{figure*}

\renewcommand{\dblfloatpagefraction}{0.7}
\begin{figure*}
  \centering
    \includegraphics[width=\linewidth]{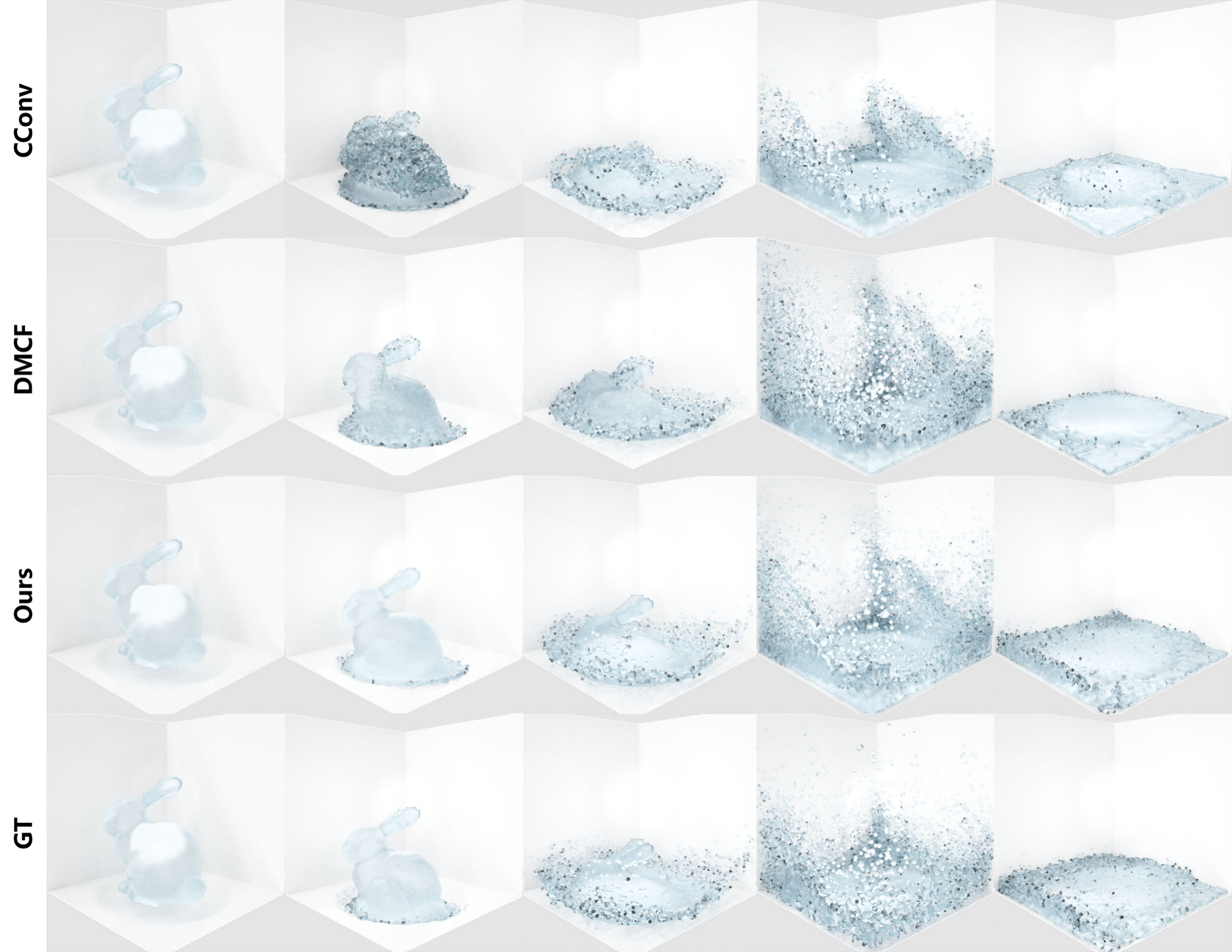}
    \caption{Evaluation of the irregular initial shapes: the Stanford Bunny fluid free-falls into a box with a circular depression at the bottom. The sequence includes an initial frame and four simulation frames. The four moments we selected include 'the first meet of surface', 'half rabbit submerged', 'splash hits maximum height', and 'water calms at the final frame'.}
    \label{fig:bunny}
\end{figure*}


\subsection{Evaluation metrics}
\label{subsec:Evaluation metrics}
To evaluate the prediction accuracy of each method, we use the chamfer distance (CD) by calculating the average positional error between the predicted positions to the ground truth. We start the assessment by using every \(5^{th}\) frame for initialization and compute the deviations from the ground truth between two consecutive frames, denoted as \(t+1\) and \(t+2\). Furthermore, we measure long-term similarity by reporting the average distance from the ground-truth particles to the closest particle in the prediction for the whole sequence of \(n\) frames. This distance for frame \(n\) is computed as follows:
\begin{equation}\label{eq11}
    \begin{aligned}
d^{n} =\frac{1}{N} \sum_{i=1}^{N} \min_{x^n\in X^n}\left \| \hat{x}_i^n -x^n \right \|  _2.
    \end{aligned}
\end{equation}
Where \(X^n\) represents the set of predicted particle positions for frame \(n\), \(\hat{x}_i^n\) corresponds to the ground-truth position for particle \(i\), and \(N\) denotes the total number of particles.


We quantify the probability distribution divergence between the prediction and ground truth by calculating the Wasserstein distance \cite{rubner2000earth}. The Wasserstein distance, also known as the Earth Mover's Distance (EMD), is a measure of the transportation cost required to transform one probability distribution into another, which can be used to measure the similarity between two distributions. 

Notably, the chamfer distance (Equation \ref{eq11}) effectively calculates the aggregate simulation error at the particle level,  thereby it can serve as a metric of the overall control capability in fluid simulations. However, it might not adequately capture global shape discrepancies, especially when there are substantial deformations in the shapes (for instance, in fluid motion). We find that the Wasserstein distance, which measures the error in the overall probability distribution, serves as a more indicative metric of adherence to physical laws. To show our network can not only achieve precise global fluid control but also ensure adherence to physical law constraints, it must excel in both metrics compared to previous methods.

Maximum density is defined as the highest density value among all fluid particles. This metric correlates with the fluid's incompressibility and provides an indication of a method's stability. The max density error is calculated as follows:
\begin{equation}
     e =\left | 1-\frac{\max_{i} \rho (x_i)}{ \max_{i} \rho ( \hat{x}_i)}  \right | 
\end{equation}
Here, \(\rho\) represents the function for calculating the density of particle \(i\), where \(x_i\) denotes the predicted particle \(i\) and \(\hat{x}_i^n\) corresponds to the ground-truth particle \(i\).

\subsection{Comparative Experiments}
\subsubsection{Quantitative Evaluation}
Table \ref{tab:Quantitative Experiments} presents comparative experiments conducted on the Liquid(6k) and DamBreak datasets. \textit{Average pos error} and \textit{Average distance of \(n\) frames sequence} assess error at the particle level, with the former evaluating short-term fluid prediction (next two frames) and the latter assessing long-term prediction (n frames). \textit{Wasserstein distance} evaluates prediction accuracy at the level of probability distribution, while \textit{Max density error} calculates the error between the maximum density of predicted particles and the ground truth, which correlates with the fluid's incompressibility and provides an indication of a method's stability. Finally, \textit{Frame inference time} measures the average speed of inference per frame.

DFSPH \cite{bender2015divergence} is a traditional fluid simulation method, which exhibits accuracy close to ground truth. However, the computational demands of traditional methods are hundreds of times higher than data-driven methods. We compare the simulation results with both traditional machine learning \cite{ladicky2015data} and deep learning approaches \cite{li2018learning, ummenhofer2019lagrangian, prantl2022guaranteed, thomas2019kpconv, Wang_2018_CVPR}. Traditional machine learning methods, exemplified by regression forests \cite{ladicky2015data}, boast rapid inference speeds. However, they exhibit significant shortcomings in stability and generalizability compared to other existing approaches, which will be further analyzed in Section \ref{complex capability}. It is worth noting that, despite DMCF \cite{prantl2022guaranteed} being an improved version of CConv \cite{ummenhofer2019lagrangian}, it underperforms CConv in terms of the chamfer distance metric. This is because the anti-symmetric continuous convolution layer introduces a strong constraint, which makes the problem considerably more challenging for the network to solve. Merely appending the ASCC at the end in \cite{prantl2022guaranteed} made it more complex to learn and fit the global fluid control. As a trade-off, the network exhibits improved adherence to physical laws and enhanced generalization capabilities: the Wasserstein distance to the ground truth has also shown a decrease, indicating a distribution that more closely approximates the physically accurate ground truth. As for our network, while it may not be the fastest neural network, it outperforms in both prediction accuracy and probability distribution divergence compared to previous models, maintaining a small error of max density. A smaller Max density error signifies enhanced fluid incompressibility and stability. These achievements are attributed to the design of our Dual-pipeline architecture, which attains equilibrium between global fluid control and adherence to physical law constraints.

\subsubsection{Qualitative Evaluation}
\label{subsubsection:Qualitative Evaluation}
Figure \ref{fig:Qualitative Evaluation} presents a visual comparison of our network with the most stable methods CConv and DMCF on the Liquid3D(complex) dataset. To better analyze the underlying physical principles during the dynamic simulation process, we selected five frames from the free-fall to collision with the container's bottom for analysis. The visual contrast may not be immediately striking at first glance, but the details inherently reveal the level of commitment to following physical laws. In detail, CConv starts to deform its structure from the second frame onwards, which significantly deviates from the expected physical laws of free fall. This is attributed to CConv's lack of emphasis on the law of momentum conservation, allowing it to overfit gravity and negate it. This reduces the problem to a local per-particle problem. In contrast, DMCF exhibits an improved adherence to the laws of physics during the free fall phase, thanks to the inclusion of the antisymmetric design. However, DMCF's approach of simply adding ASCC as the final output layer compromises the overall stability of the fluid. This instability becomes evident from the second frame, where the edges of the cube exhibit noticeable distortion. Our network demonstrates an accurate simulation by providing a stable global context while maintaining a strong grasp of physical principles throughout the simulation process, bringing it closer to the ground truth.

Figure \ref{fig:bunny} shows the results of the evaluation for irregular initial shapes (Stanford Bunny), which is completely unseen, as the initial shapes in the Liquid3D dataset are all regular. We selected five moments for comparison. In the face of complex initial fluid shapes, CConv, lacking a grasp of physical principles, resulted in the rabbit completely collapsing and deforming during its fall; On the other hand, DMCF more accurately reflected the physical properties of free fall but showed inadequacy in global fluid control. This is evident as some liquid particles splashed out of the container during the fall and collision, and the liquid reached a noticeably higher and more chaotic state than the ground truth at the peak moment. Our network, however, closely aligns with the ground truth throughout the free fall and collision, demonstrating its exceptional generalization capabilities in complex and unseen initial shapes.

\begin{figure*}
  \centering
    \includegraphics[width=\linewidth]{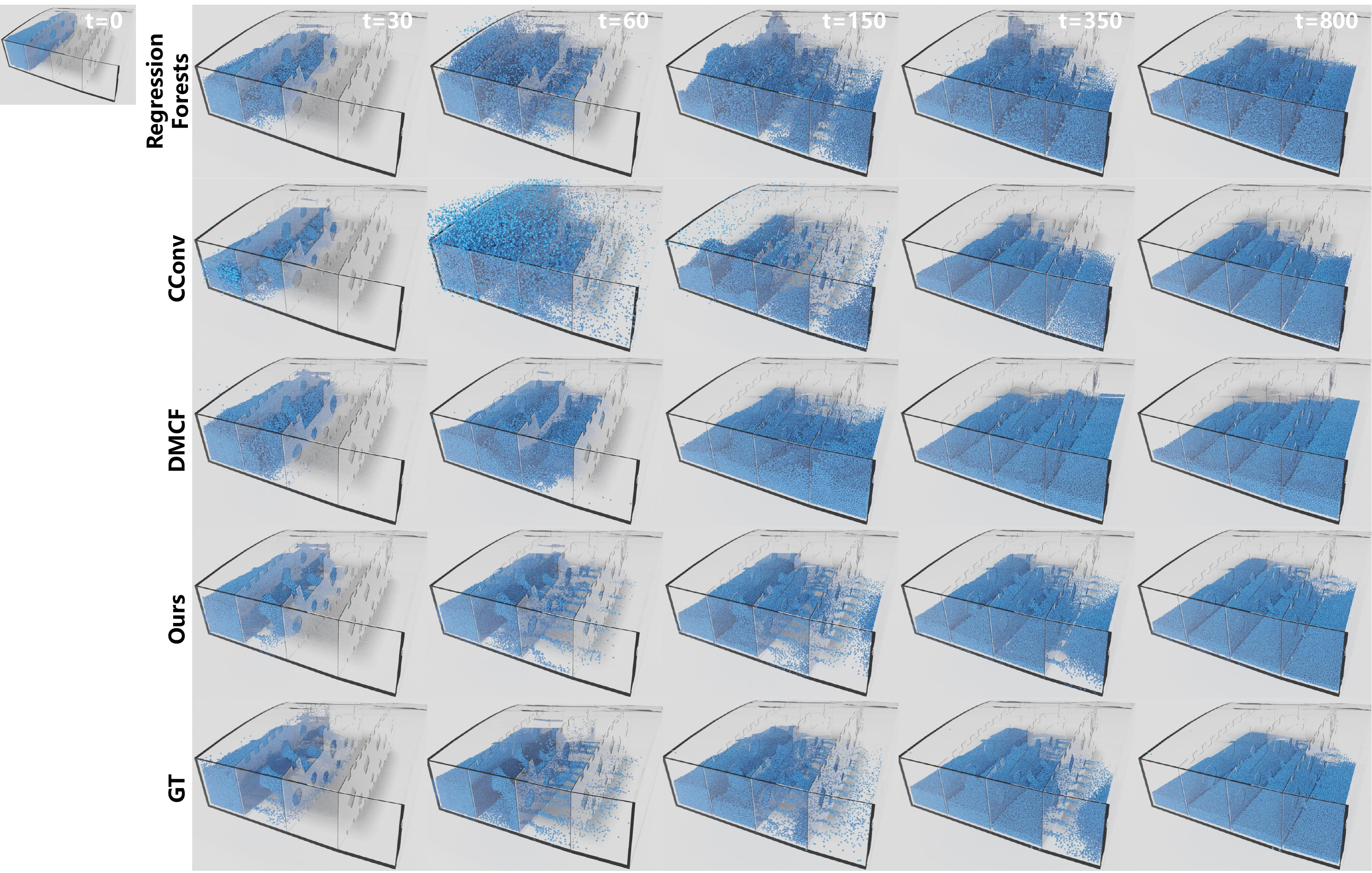}
    \caption{Qualitative Evaluation on Tank3D Dataset. Comparative experiments in complex tank scenarios provide a deeper exploration of models' capability to handle complex scenarios, emphasizing adherence to physical laws, fluid-solid interaction proficiency, and generalization capacity. The image in the top left corner represents the initial state at frame 0.}
    \label{fig:tankdataset}
\end{figure*}
\begin{figure*}
  \centering
    \includegraphics[width=\linewidth]{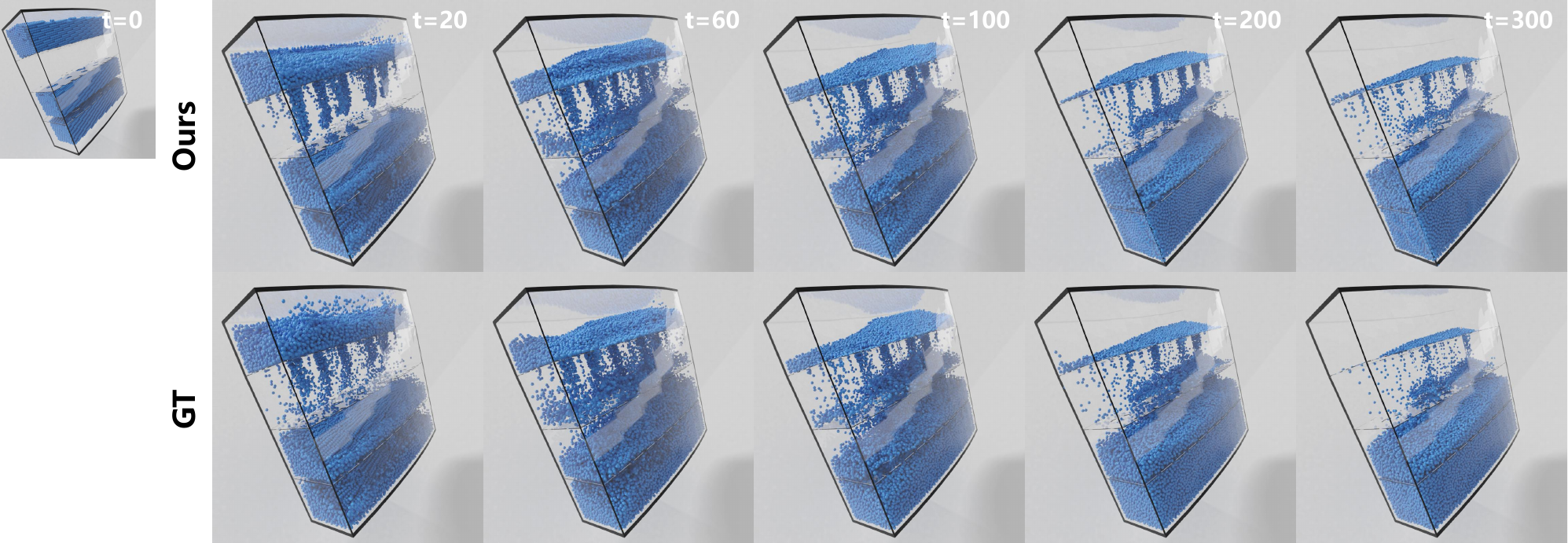}
    \caption{Evaluation on our method's rotation invariance. Our approach maintains accurate predictions even for scenes that are randomly rotated and unseen before.}
    \label{fig:tankdataset_incline}
\end{figure*}

\begin{table*}
    \renewcommand{\arraystretch}{0.85}
  \centering
\resizebox{1.0\linewidth}{!}{
\begin{tabular}{lccccc}
    \toprule[1pt]
    \setlength{\tabcolsep}{\linewidth}
{\multirow{2}{*}{Method}} & \multicolumn{2}{c}{Average pos error(mm)} & Average distance of & Wasserstein & Max density \\ \cline{2-3}
                        & t+1            & {t+2} & \(n\) frames sequence \(d^n\)(mm) & distance(mm)    & error(\(g/cm^{3}\))                                       \\ \specialrule{0em}{0pt}{1pt}     \midrule \midrule
            Main Path (w/o TaIM)   &          0.62            &        1.80        &   30.65  &      0.26  & 0.12                                \\
          Main Path        &       0.58              &      1.61        &  30.67    &   0.23  &  0.11\\
           Constraint-guided Path (w/o TaIM)          &       0.75              &     2.18        &36.94  &     0.30    & 0.15      \\ 
           
           Constraint-guided Path          &       0.67               &      1.86          & 32.20   &      0.25 & 0.12                                               \\ 
          Ours (w/o TaIM)           &      0.57           &      1.59        &  30.04    &      0.20   &0.06                               \\ 
           Ours          &       \textbf{0.52}               &      \textbf{1.45}          & \textbf{29.42}   &      \textbf{0.19}  & \textbf{0.06}   \\ 
     \bottomrule[1pt]
    \specialrule{0em}{0pt}{1pt}
\end{tabular}}
  \caption{Ablation Study. Our ablation experiments were conducted on the 'Liquid (complex)' dataset, testing the following configurations: a single Main Path with and without TaIM (\textbf{T}ype-\textbf{a}ware \textbf{I}nput \textbf{M}odule), a single Constraint-guided Path with and without TaIM, and our final Dual-pipeline network with and without TaIM.}
  \label{tab:Ablation Study}
\end{table*}

\begin{table*}
\small
  \renewcommand{\arraystretch}{0.6}
  \centering
  \resizebox{1.0\linewidth}{!}{
\scriptsize
\begin{tabular}{lccccc}
    \toprule[0.6pt]
{\multirow{2}{*}{Method}} & \multicolumn{2}{c}{Average pos error(mm)} & Average distance of & Wasserstein  & Max density  \\ \cline{2-3}
                        & t+1            & {t+2} & \(n\) frames sequence \(d^n\)(mm) &   distance(mm)  & error(\(g/cm^{3}\))                                          \\ \specialrule{0em}{0pt}{1pt}     \midrule \midrule
                 CConv     &          0.66            &        1.86        &   30.83  &       0.26      & 0.12            \\
          CConv(w/ TaIM)      &      0.59           &      1.69        &  30.70    &      0.23     & 0.11                                          \\
           DMCF           &       0.73              &      2.08         & 47.89    &     0.22    &0.07            \\ 
           
           DMCF(w/ TaIM)           &    0.69               &       2.03      &  46.58  &    0.22     &0.07                                                 \\ 
           Ours (w/o TaIM)           &  0.57                    &     1.59       & 30.04  &  0.20 & 0.06  \\ 
           Ours          &    \textbf{0.52}                  &  \textbf{1.45}          &  \textbf{29.42} &  \textbf{0.19} & \textbf{0.06}                                         \\
     \bottomrule[0.6pt]
\end{tabular}}
  \caption{The plug-and-play usability of TaIM. We incorporated TaIM as a preprocessing step before input in CConv, DMCF, and our network, and compared the performance differences with and without TaIM. We can easily integrate TaIM into networks with similar convolution kernels, resulting in a significant performance improvement. This demonstrates that the concept behind TaIM is beneficial for fluid-solid coupling.}
  \label{tab:Plug-and-play}
\end{table*}
\subsubsection{Capability of Handling Complex Scene} \label{complex capability}
In the experimental results presented above, we employ the same experimental methods and datasets as existing approaches, investigating the fundamental capabilities of our proposed method and existing methods in fluid simulation tasks. To further explore the network's ability to handle complicated scenes, we propose a dataset, Tank3D, inspired by the Liquid3D dataset. Although smaller in scale, it features significantly greater complexity.

\begin{figure*}
  \centering
    \includegraphics[width=\linewidth]{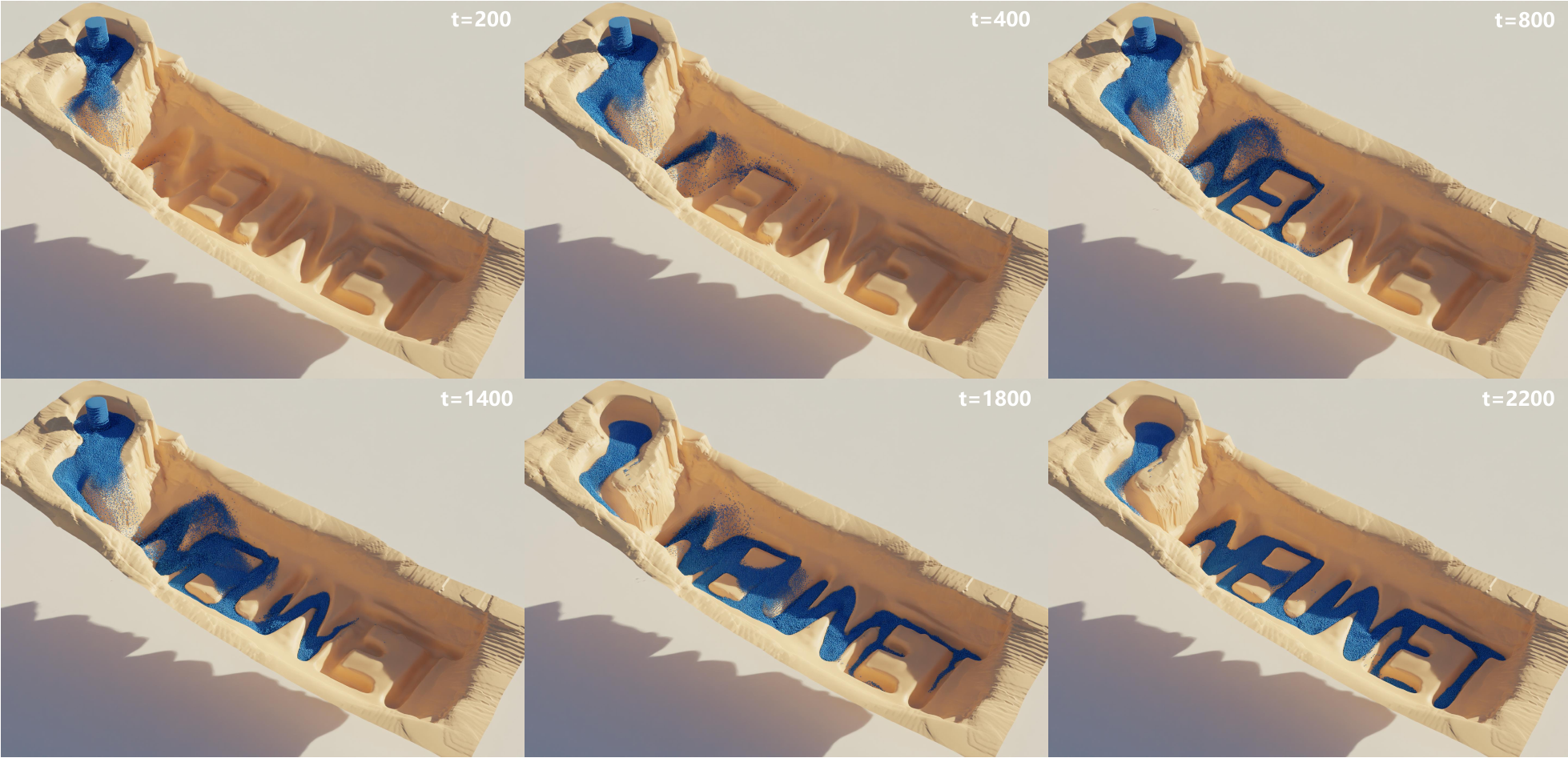}
    \caption{Generalization to complex terrains and varying fluid particle scenarios: fluid emitted from a particle emitter flows from a higher elevation in a complex valley, eventually filling the depressions in a plain. The video is available in our GitHub repository.}
    \label{fig:canyon}
\end{figure*}
We train the model for 50,000 iterations using the Tank3D dataset, followed by qualitative comparisons as shown in Figure \ref{fig:tankdataset}. Regression forests \cite{ladicky2015data} stand as a classic example of traditional machine learning methods applied to fluid simulation tasks. Like many machine learning techniques, they struggle to generalize well to scenarios not encountered in the training set. Another practical problem is that water surfaces often remain unsettled and fail to become perfectly flat, even by the final frame. CConv \cite{ummenhofer2019lagrangian} lacks guidance from physical principles, leading to observable severe compression of the fluid in the first image, which significantly deviates from the incompressibility of fluids. Due to the kernel computation method of the SPH approach, intense mutual compression among fluid particles results in a drastic increase in the velocity calculations for each particle, causing the dispersion of particles seen in the second image. The performance of DMCF \cite{prantl2022guaranteed} is considerably more stable. However, it still demonstrates a deficiency in capturing the interactions between fluids and complex baffles within the tank, manifesting as the fluid disregarding the presence of baffles and flowing directly downwards. In contrast, our method yields simulation results that closely align with reality, both in terms of overall fluid control and adherence to physical laws.

Furthermore, Figure \ref{fig:tankdataset_incline} demonstrates the rotation invariance of our method. Faced with scenes that have undergone random rotations and were not encountered in the training set, our method still produces accurate predictions. This indicates that our approach has genuinely learned the principles of fluid dynamics and adheres well to the physical laws of the real world. This is difficult to achieve for other methods, especially traditional machine learning approaches.

\subsection{Ablation Study}
We conduct an ablation study to assess the design choices of our network, shown in Table \ref{tab:Ablation Study}. This study employs the same evaluation metrics as the comparative experiments. We systematically analyze the contributions of each pathway independently and evaluate the enhancements provided by the Type-aware Input Module to both pathways. In addition, we also confirm the plug-and-play capability of the Type-aware Input Module. We conducted separate tests by adding this module to CConv, DMCF, and our network, as they all utilize similar convolution kernels. As illustrated in Table \ref{tab:Plug-and-play}, this module proves to be instrumental in enhancing fluid simulation accuracy, indicating its pivotal role in achieving fluid-solid coupling.

\subsection{Generalization}
We validate the generalization capability of our network to an unseen and complex scenario, where fluid, emitted from a particle emitter, flows from a high point in a complex valley terrain down to fill a depression in a plain, demonstrating the network's generalization ability regarding scene geometry and the number of particles, as shown in Figure \ref{fig:canyon}. The results demonstrate the strong generalization capabilities of our method in handling complex fluid shapes and terrains. This is attributed to our Dual-pipeline architecture, which effectively learns fluid dynamics while tightly adhering to physical laws.

\section{Conclusion}
\label{sec:conclusion}
We have designed an Attention-based Dual-pipeline network architecture, which provides stable global fluid control and ensures adherence to physical law constraints in fluid simulation. The Dual-pipeline feature fusion architecture also holds the potential to inspire work in other domains. Additionally, we propose a Type-aware Input Module, which plays a pivotal role in facilitating the fluid-solid particle coupling. Our network achieves not merely a quantitative improvement in performance metrics, but a qualitative leap in neural network-based simulation fidelity adhering to the laws of the physical world, enhancing the potential for real-time simulation generalization across diverse fluid scenarios, particularly in scenarios such as interactive gaming environments, and virtual reality simulations. However, there still remains untapped potential in enabling neural networks to learn real-world physics. We will make our code publicly available to contribute to the ongoing research.

\section*{Acknowledgments}
The work reported in this paper is supported by the National Natural Science Foundation of China [52375514].

\section*{CRediT authorship contribution statement}
\textbf{Yu Chen:} Methodology, Algorithm, Validation, Writing - original draft. \textbf{Shuai Zheng:} Conceptualization, Funding acquisition, Project administration, Supervision, Writing - review \& editing. \textbf{Menglong Jin:} Visualization, Data curation. \textbf{Yan Chang:} Visualization. \textbf{Nianyi Wang:} Investigation, Writing - review \& editing.

\section*{Declaration of competing interest}
The authors declare that they have no known competing financial interests or personal relationships that could have appeared to influence the work reported in this paper.

\section*{Data availability}
We have shared the link to our code and data on the following GitHub repository: \href{https://github.com/chenyu-xjtu/DualFluidNet}{https://github.com/chenyu-xjtu/DualFluidNet}. 





\bibliographystyle{plain}
\bibliography{main}

\begin{thebibliography}{10}

\bibitem{battaglia2016interaction}
Peter Battaglia, Razvan Pascanu, Matthew Lai, Danilo Jimenez~Rezende, et~al.
\newblock Interaction networks for learning about objects, relations and physics.
\newblock {\em Advances in neural information processing systems}, 29, 2016.

\bibitem{bender2015divergence}
Jan Bender and Dan Koschier.
\newblock Divergence-free smoothed particle hydrodynamics.
\newblock In {\em Proceedings of the 14th ACM SIGGRAPH/Eurographics symposium on computer animation}, pages 147--155, 2015.

\bibitem{cai2021physics}
Shengze Cai, Zhiping Mao, Zhicheng Wang, Minglang Yin, and George~Em Karniadakis.
\newblock Physics-informed neural networks (pinns) for fluid mechanics: A review.
\newblock {\em Acta Mechanica Sinica}, 37(12):1727--1738, 2021.

\bibitem{chen2023rotation}
Yu~Chen and Pengcheng Shi.
\newblock Rotation-invariant completion network.
\newblock {\em arXiv preprint arXiv:2308.11979}, 2023.

\bibitem{dai2021attentional}
Yimian Dai, Fabian Gieseke, Stefan Oehmcke, Yiquan Wu, and Kobus Barnard.
\newblock Attentional feature fusion.
\newblock In {\em Proceedings of the IEEE/CVF winter conference on applications of computer vision}, pages 3560--3569, 2021.

\bibitem{DENG2024106085}
Yong Deng, Haifeng Wang, and Xianming Shi.
\newblock Physics-guided neural network for predicting asphalt mixture rutting with balanced accuracy, stability and rationality.
\newblock {\em Neural Networks}, 172:106085, 2024.

\bibitem{LSTM1}
Carmina Fjellstr{\"o}m.
\newblock Long short-term memory neural network for financial time series.
\newblock {\em 2022 IEEE International Conference on Big Data (Big Data)}, pages 3496--3504, 2022.

\bibitem{GAO202082}
Zhifan Gao, Xin Wang, Shanhui Sun, Dan Wu, Junjie Bai, Youbing Yin, Xin Liu, Heye Zhang, and Victor Hugo~C. {de Albuquerque}.
\newblock Learning physical properties in complex visual scenes: An intelligent machine for perceiving blood flow dynamics from static ct angiography imaging.
\newblock {\em Neural Networks}, 123:82--93, 2020.

\bibitem{gingold1977smoothed}
Robert~A Gingold and Joseph~J Monaghan.
\newblock Smoothed particle hydrodynamics: theory and application to non-spherical stars.
\newblock {\em Monthly notices of the royal astronomical society}, 181(3):375--389, 1977.

\bibitem{griepentrog2008bi}
Jens~Andr{\'e} Griepentrog, Wolfgang H{\"o}ppner, Hans-Christoph Kaiser, and Joachim Rehberg.
\newblock A bi-lipschitz continuous, volume preserving map from the unit ball onto a cube.
\newblock {\em Note di Matematica}, 28(1):177--193, 2008.

\bibitem{guan2022neurofluid}
Shanyan Guan, Huayu Deng, Yunbo Wang, and Xiaokang Yang.
\newblock Neurofluid: Fluid dynamics grounding with particle-driven neural radiance fields.
\newblock In {\em International Conference on Machine Learning}, pages 7919--7929. PMLR, 2022.

\bibitem{Context-Aware}
Yanbu Guo, Dongming Zhou, Pu~Li, Chaoyang Li, and Jinde Cao.
\newblock Context-aware poly(a) signal prediction model via deep spatial–temporal neural networks.
\newblock {\em IEEE Transactions on Neural Networks and Learning Systems}, pages 1--13, 2022.

\bibitem{Variational-gated}
Yanbu Guo, Dongming Zhou, Xiaoli Ruan, and Jinde Cao.
\newblock Variational gated autoencoder-based feature extraction model for inferring disease-mirna associations based on multiview features.
\newblock {\em Neural Networks}, 165:491--505, 2023.

\bibitem{he2016deep}
Kaiming He, Xiangyu Zhang, Shaoqing Ren, and Jian Sun.
\newblock Deep residual learning for image recognition.
\newblock In {\em Proceedings of the IEEE conference on computer vision and pattern recognition}, pages 770--778, 2016.

\bibitem{hermosilla2018monte}
Pedro Hermosilla, Tobias Ritschel, Pere-Pau V{\'a}zquez, {\`A}lvar Vinacua, and Timo Ropinski.
\newblock Monte carlo convolution for learning on non-uniformly sampled point clouds.
\newblock {\em ACM Transactions on Graphics (TOG)}, 37(6):1--12, 2018.

\bibitem{JIN2021109951}
Xiaowei Jin, Shengze Cai, Hui Li, and George~Em Karniadakis.
\newblock Nsfnets (navier-stokes flow nets): Physics-informed neural networks for the incompressible navier-stokes equations.
\newblock {\em Journal of Computational Physics}, 426:109951, 2021.

\bibitem{kakuda2021data}
Kazuhiko Kakuda, Yuto Morimasa, Tomoyuki Enomoto, Wataru Okaniwa, and Shinichiro Miura.
\newblock Data-driven fluid flow simulations by using convolutional neural network.
\newblock In {\em Computational and Experimental Simulations in Engineering: Proceedings of ICCES 2020. Volume 1 26}, pages 14--19. Springer, 2021.

\bibitem{KASHEFI202380}
Ali Kashefi and Tapan Mukerji.
\newblock Prediction of fluid flow in porous media by sparse observations and physics-informed pointnet.
\newblock {\em Neural Networks}, 167:80--91, 2023.

\bibitem{koschier2020smoothed}
Dan Koschier, Jan Bender, Barbara Solenthaler, and Matthias Teschner.
\newblock Smoothed particle hydrodynamics techniques for the physics based simulation of fluids and solids.
\newblock {\em arXiv preprint arXiv:2009.06944}, 2020.

\bibitem{ladicky2015data}
L'ubor Ladick{\`y}, SoHyeon Jeong, Barbara Solenthaler, Marc Pollefeys, and Markus Gross.
\newblock Data-driven fluid simulations using regression forests.
\newblock {\em ACM Transactions on Graphics (TOG)}, 34(6):1--9, 2015.

\bibitem{li2018multidisciplinary}
Baotong Li, Honglei Liu, and Shuai Zheng.
\newblock Multidisciplinary topology optimization for reduction of sloshing in aircraft fuel tanks based on sph simulation.
\newblock {\em Structural and multidisciplinary optimization}, 58:1719--1736, 2018.

\bibitem{Learning-spatiotemporal}
Weihua Li, Yanbu Guo, Bingyi Wang, and Bei Yang.
\newblock Learning spatiotemporal embedding with gated convolutional recurrent networks for translation initiation site prediction.
\newblock {\em Pattern Recognition}, 136:109234, 2023.

\bibitem{li2018learning}
Yunzhu Li, Jiajun Wu, Russ Tedrake, Joshua~B Tenenbaum, and Antonio Torralba.
\newblock Learning particle dynamics for manipulating rigid bodies, deformable objects, and fluids.
\newblock {\em arXiv preprint arXiv:1810.01566}, 2018.

\bibitem{ling2016reynolds}
Julia Ling, Andrew Kurzawski, and Jeremy Templeton.
\newblock Reynolds averaged turbulence modelling using deep neural networks with embedded invariance.
\newblock {\em Journal of Fluid Mechanics}, 807:155--166, 2016.

\bibitem{liu2023sfusion}
Zecheng Liu, Jia Wei, Rui Li, and Jianlong Zhou.
\newblock Sfusion: Self-attention based n-to-one multimodal fusion block.
\newblock In {\em International Conference on Medical Image Computing and Computer-Assisted Intervention}, pages 159--169. Springer, 2023.

\bibitem{macklin2013position}
Miles Macklin and Matthias M{\"u}ller.
\newblock Position based fluids.
\newblock {\em ACM Transactions on Graphics (TOG)}, 32(4):1--12, 2013.

\bibitem{morton2018deep}
Jeremy Morton, Antony Jameson, Mykel~J Kochenderfer, and Freddie Witherden.
\newblock Deep dynamical modeling and control of unsteady fluid flows.
\newblock {\em Advances in Neural Information Processing Systems}, 31, 2018.

\bibitem{mrowca2018flexible}
Damian Mrowca, Chengxu Zhuang, Elias Wang, Nick Haber, Li~F Fei-Fei, Josh Tenenbaum, and Daniel~L Yamins.
\newblock Flexible neural representation for physics prediction.
\newblock {\em Advances in neural information processing systems}, 31, 2018.

\bibitem{prantl2022guaranteed}
Lukas Prantl, Benjamin Ummenhofer, Vladlen Koltun, and Nils Thuerey.
\newblock Guaranteed conservation of momentum for learning particle-based fluid dynamics.
\newblock {\em Advances in Neural Information Processing Systems}, 35:6901--6913, 2022.

\bibitem{qi2017pointnet}
Charles~R Qi, Hao Su, Kaichun Mo, and Leonidas~J Guibas.
\newblock Pointnet: Deep learning on point sets for 3d classification and segmentation.
\newblock In {\em Proceedings of the IEEE conference on computer vision and pattern recognition}, pages 652--660, 2017.

\bibitem{qi2017pointnet++}
Charles~Ruizhongtai Qi, Li~Yi, Hao Su, and Leonidas~J Guibas.
\newblock Pointnet++: Deep hierarchical feature learning on point sets in a metric space.
\newblock {\em Advances in neural information processing systems}, 30, 2017.

\bibitem{rubner2000earth}
Yossi Rubner, Carlo Tomasi, and Leonidas~J Guibas.
\newblock The earth mover's distance as a metric for image retrieval.
\newblock {\em International journal of computer vision}, 40:99--121, 2000.

\bibitem{SAHA2021359}
Priyabrata Saha, Saurabh Dash, and Saibal Mukhopadhyay.
\newblock Physics-incorporated convolutional recurrent neural networks for source identification and forecasting of dynamical systems.
\newblock {\em Neural Networks}, 144:359--371, 2021.

\bibitem{sanchez2020learning}
Alvaro Sanchez-Gonzalez, Jonathan Godwin, Tobias Pfaff, Rex Ying, Jure Leskovec, and Peter Battaglia.
\newblock Learning to simulate complex physics with graph networks.
\newblock In {\em International conference on machine learning}, pages 8459--8468. PMLR, 2020.

\bibitem{shao2022transformer}
Yidi Shao, Chen~Change Loy, and Bo~Dai.
\newblock Transformer with implicit edges for particle-based physics simulation.
\newblock In {\em European Conference on Computer Vision}, pages 549--564. Springer, 2022.

\bibitem{solenthaler2009predictive}
Barbara Solenthaler and Renato Pajarola.
\newblock Predictive-corrective incompressible sph.
\newblock In {\em ACM SIGGRAPH 2009 papers}, pages 1--6. 2009.

\bibitem{thomas2019kpconv}
Hugues Thomas, Charles~R Qi, Jean-Emmanuel Deschaud, Beatriz Marcotegui, Fran{\c{c}}ois Goulette, and Leonidas~J Guibas.
\newblock Kpconv: Flexible and deformable convolution for point clouds.
\newblock In {\em Proceedings of the IEEE/CVF international conference on computer vision}, pages 6411--6420, 2019.

\bibitem{tompson2017accelerating}
Jonathan Tompson, Kristofer Schlachter, Pablo Sprechmann, and Ken Perlin.
\newblock Accelerating eulerian fluid simulation with convolutional networks.
\newblock In {\em International Conference on Machine Learning}, pages 3424--3433. PMLR, 2017.

\bibitem{ummenhofer2019lagrangian}
Benjamin Ummenhofer, Lukas Prantl, Nils Thuerey, and Vladlen Koltun.
\newblock Lagrangian fluid simulation with continuous convolutions.
\newblock In {\em International Conference on Learning Representations}, 2019.

\bibitem{vaswani2017attention}
Ashish Vaswani, Noam Shazeer, Niki Parmar, Jakob Uszkoreit, Llion Jones, Aidan~N Gomez, {\L}ukasz Kaiser, and Illia Polosukhin.
\newblock Attention is all you need.
\newblock {\em Advances in neural information processing systems}, 30, 2017.

\bibitem{Fusion1}
Changcheng Wang, Dongming Zhou, Yongsheng Zang, Rencan Nie, and Yanbu Guo.
\newblock A deep and supervised atrous convolutional model for multi-focus image fusion.
\newblock {\em IEEE Sensors Journal}, 21(20):23069--23084, 2021.

\bibitem{Fusion-Based}
Kaili Wang and Min Li.
\newblock Fusion-based deep learning architecture for detecting drug-target binding affinity using target and drug sequence and structure.
\newblock {\em IEEE Journal of Biomedical and Health Informatics}, 27(12):6112--6120, 2023.

\bibitem{Wang_2018_CVPR}
Shenlong Wang, Simon Suo, Wei-Chiu Ma, Andrei Pokrovsky, and Raquel Urtasun.
\newblock Deep parametric continuous convolutional neural networks.
\newblock In {\em Proceedings of the IEEE Conference on Computer Vision and Pattern Recognition (CVPR)}, June 2018.

\bibitem{wang2019dynamic}
Yue Wang, Yongbin Sun, Ziwei Liu, Sanjay~E Sarma, Michael~M Bronstein, and Justin~M Solomon.
\newblock Dynamic graph cnn for learning on point clouds.
\newblock {\em ACM Transactions on Graphics (tog)}, 38(5):1--12, 2019.

\bibitem{woodward2023physics}
Michael Woodward, Yifeng Tian, Criston Hyett, Chris Fryer, Mikhail Stepanov, Daniel Livescu, and Michael Chertkov.
\newblock Physics-informed machine learning with smoothed particle hydrodynamics: Hierarchy of reduced lagrangian models of turbulence.
\newblock {\em Physical Review Fluids}, 8(5):054602, 2023.

\bibitem{ye2019smoothed}
Ting Ye, Dingyi Pan, Can Huang, and Moubin Liu.
\newblock Smoothed particle hydrodynamics (sph) for complex fluid flows: Recent developments in methodology and applications.
\newblock {\em Physics of Fluids}, 31(1), 2019.

\bibitem{Underwater}
Weidong Zhang, Ling Zhou, Peixian Zhuang, Guohou Li, Xipeng Pan, Wenyi Zhao, and Chongyi Li.
\newblock Underwater image enhancement via weighted wavelet visual perception fusion.
\newblock {\em IEEE Transactions on Circuits and Systems for Video Technology}, pages 1--1, 2023.

\bibitem{zhao2023cddfuse}
Zixiang Zhao, Haowen Bai, Jiangshe Zhang, Yulun Zhang, Shuang Xu, Zudi Lin, Radu Timofte, and Luc Van~Gool.
\newblock Cddfuse: Correlation-driven dual-branch feature decomposition for multi-modality image fusion.
\newblock In {\em Proceedings of the IEEE/CVF Conference on Computer Vision and Pattern Recognition}, pages 5906--5916, 2023.

\bibitem{zheng2021topology}
Shuai Zheng, Fan Gao, Ziyu Zhang, Honglei Liu, and Baotong Li.
\newblock Topology optimization on fuel tank rib structures for fuel sloshing suppression based on hybrid fluid--solid sph simulation.
\newblock {\em Thin-Walled Structures}, 165:107938, 2021.

\end{thebibliography}

\end{document}